\documentclass[review]{elsarticle}
\usepackage{graphicx}
\usepackage{booktabs}
\usepackage{threeparttable}
\usepackage{indentfirst}
\usepackage{url}
\usepackage{lineno,hyperref}
\usepackage{color}
\begin{document}

\begin{frontmatter}

\title{Identifying Emotion from Natural Walking}
%\tnotetext[mytitlenote]{Fully documented templates are available in the elsarticle package on \href{http://www.ctan.org/tex-archive/macros/latex/contrib/elsarticle}{CTAN}.}

%% Group authors per affiliation:
\author{Liqing Cui$^\dag$$^\ddag$, Shun Li$^\dag$$^\ddag$, Wan Zhang$^\dag$$^\ddag$, Zhan Zhang$^\dag$, Tingshao Zhu$^\dag$\corref{mycorrespondingauthor}}
\address{$^\dag$ Institute of Psychology,Chinese Academy of Sciences}
\address{$^\ddag$ The 6th Research Institute of China Electronics Corporation(National Computer System Engineering Research Institute of China)}

%% or include affiliations in footnotes:
%\author[mymainaddress,mysecondaryaddress]{Elsevier Inc}
%\ead[url]{www.elsevier.com}

%\author[mysecondaryaddress]{Global Customer Service\corref{mycorrespondingauthor}}
\cortext[mycorrespondingauthor]{Corresponding author}
%\ead{support@elsevier.com}

%\address[mymainaddress]{1600 John F Kennedy Boulevard, Philadelphia}
%\address[mysecondaryaddress]{360 Park Avenue South, New York}

\begin{abstract}
\noindent
%【在摘要的前面一两句话，简要介绍一下步态识别情绪的研究意义。】
Emotion identification from gait aims to automatically determine person’s  affective state, it has attracted a great deal of interests and offered immense potential value in  action tendency, health care, psychological detection and human-computer(robot) interaction.
In this paper,
we propose a new method of identifying emotion from natural walking,
and analyze the relevance between the traits of walking and affective states.
%Considering the redundant information exist in high dimension data,
After obtaining the pure acceleration data of wrist and ankle,
we set a moving average filter window with different sizes $w$, then extract $114$ features including time-domain, frequency-domain, power and distribution features from each data slice, and run principal component analysis (PCA) to reduce dimension.
In experiments, we train SVM, Decision Tree, multilayerperception, Random Tree and Random Forest classification models,
and compare the classification accuracy on data of wrist and ankle with respect to different $w$.
The performance of emotion identification on acceleration data of ankle is better than wrist.
Comparing different classification models' results,
SVM has best accuracy of identifying anger and happy could achieve $90.31\%$ and $89.76\%$ respectively, and identification ratio of anger-happy is $87.10\%$.
The anger-neutral-happy classification reaches $85\%$-$78\%$-$78\%$.
The results show that it is capable of identifying personal emotional states through the gait of walking.

\end{abstract}

\begin{keyword}
\texttt{Seonsor mining, emotion identification, cellphone, accelerometer sensor}
%\MSC[2010] 00-01\sep  99-00??
\end{keyword}

\end{frontmatter}

%\linenumbers

\section{Introduction}
Nonverbal communication plays a major role in future robotics.
Nonverbal signals of humans are observed to deliver additional cues for a person's physiological and psychological state and intentions,
which can be used to improve human-machine interaction and detect human's health state.
It is very challenging task to identify person's affect state automatically.
Traditionally,
emotion detection and identification is based on facial expressions,
linguistic and acoustic features in speech. Psychological studies on visual analysis of body movement show that human movement differs from other movements because it is the only visual stimulus we have experience of both perceiving and producing\cite{bealeaffect}\cite{Z.Zeng2009}.
In this paper, we propose a method of identifying human's emotion from walking.
In walking, we only record the accelerometer data of person's wrist and ankle by built-in sensors of cellphone devices.

Nowadays, mobile devices have already become indispensable communication tools in people daily life,
and more and more smartphones integrate many powerful sensors, including
GPS, light sensors, direction sensors, temperature sensors, proximity sensors,
pressure sensors, acceleration sensors, and gravity sensors.
Some of them, with advantages of small size,substantial computing power and high precision, are not only complementarily help users manage their devices intelligently,
but offer new opportunities for data mining and mining applications.  %   %【这句话说的不通顺，看不懂什么意思】

In this paper,
we use acceleration sensors in smartphone to identify human emotion based on natural walking.
To collect accelerometer data,
the participants are instructed to attach cellphones to their wrist and ankle,
then walk several minutes naturally.
Due to that we record the accelerometer and gravity data of wrist and ankle by cellphone,
we can obtain pure accrlerometer data of wrist and ankle joint.
The redundant information exist in high dimension including time-domain,
frequency-domain, power, distribution features.
We utilize Principal Component Analysis (PCA)to reduce redundant information.
On the one hand, we compare classification performance of SVM, Decision Tree,
Random Forest, Multilayerperception and Random Tree.
On the other hand, for different $w$ size, we observe data preprocessing's impact on performance.

This work is significant because the learning model permits us to obtain useful knowledge about human affection in a certain gait of millions of people to some extend -just by one person's wrist and ankle data.
Our work has a wide range of applications,
including generating a daily or weekly or monthly emotion profile to detect person's affection state change during the latest period.
If it works, we may implement in smart bracelet,
these data sent and received by wireless devices can be used as medical information.
In addition,
they can be also used for personal health by offering a benignant feedback that suggests having some exercise or entertainment.

To summarize, our research has two main contributions.
\begin{itemize}
\item We obtain pure accelerometer data from wrist and ankle in daily walking, to reveal the association between one's walking activity and her or his current emotion.
\item Data preprocess, especially in eliminating burr and noise, play a significant role apart from algorithm in improving performance of human emotion identification.
\end{itemize}

The rest of this paper is organized as follows.
Section $2$ summarizes related work about identification of emotion in walking.
Description of data base and data preprocessing and feature extraction is presented in Section $3$.
Section $4$ describes the results of the trained models and the performance of trained models for experiment,
in section $5$ we discuss our methods and summary our work.
Finally, the paper ends with a conclusion in Section $6$.
%%   outline cites comparision of pca LDA KPCA %%

\section{Related Work}
Various models to categorize emotions exist in psychology.  %【这句话不通顺】
Ekman's basic emotions, which are anger, disgust, fear, sadness and surprise,
and the dimensional pleasure-arousal-dominance(PAD)model are widely used in automatic emotion recognition\cite{Ekman1986}\cite{Russell1977}.
The PAD model spans a 3-dimensional space with the independent and bipolar axes pleasure,
arousal and dominance.
An affective state is described as a point within this state space.

Early studies based on black displays showing only the joints of the body,
then observers recognize its gender or judge it is a familiar person or not in walking\cite{Cutting1977}.
In Montepare's psychological study,
He expounded that observers can also identify emotions from variations in walking gaits\cite{Montepare1987}.
Relatively, the emotions discrimination of sadness and anger are easier than pride for observers. Pollick quantified expressive arm movements in terms of velocity and acceleration suggesting
which aspects of movement are important in recognizing emotions\cite{Pollick2001}.
Crane and Gross illustrated that body's reaction of felt and recognized emotions not only depended on gesticulatory behavior,
but associated with emotion-specific changes in gait kinematics.
In their study, they extracted some activity features,
including velocity, cadence, head orientation,
shoulder and elbow range of motion, as significant parameters to identify emotions\cite{Crane2007}.

In fact,
emotion changes rapidly even in a short walking,
and body movement is also complex,
these factors effect the correct identification accuracy.
Janssen investigated the recognition for four emotional states by means of artificial neural nets.
The recognition results reaches $83.7\%$ in average based on kinematic data for person-dependent recognition.
But inter-individual recognition remains around chance level.
There is a $79\%$ correct classification of gait patterns including calming, excitatory or no music.
Karg applied different methods such as Principal Component Analysis(PCA),
Kernal PCA(KPCA) and Linear Discriminant Analysis(LDA) into kinematic parameters of person-dependent recognition and inter-individual recognition to compare results and improve accuracy rate. LDA in combination with Naive Bayes leads to an accuracy of $91\%$ for person-dependent recognition of four discrete affective states based on observation of barely a single stride\cite{Karg200901}.
In \cite{karg2009two}, a combination of two consecutive PCA and Fourier Transformation is used for data reduction. Best accurancy is achieved for Naive Bayes with $72\%$ for the four emotions sad,neutral,happy and angry during natural walking.

A general overview of analytical techniques for clinical and biomechanical gaits analysis is given in\cite{Chau2001}\cite{Chau2001102}.
It mainly refers to classification of clinical disorders,
though the methods for feature extraction can also be taken for psychological gaits analysis.
Dimension reduction techniques such as KPCA improves recognition of age in walking\cite{Wu2007}.
The performance comparison of Principal Component Analysis (PCA) and KPCA is showed in\cite{Cao2003}.
Martinez and Kak showed that PCA can outperform when size of training sets are small to feature dimension\cite{Martinez2001}.\\
\indent Focus of our work is to extract relevant time domain, frequency domain, power, distribution features from kinematic data set to identify human emotion.
Here, we assume the arousal from different video themes is easier to identify and thus expressed in walking.
We utilize pure accelerometer data from wrist and ankle respectively to identify human emotion, and compare emotion identification accuracy of different moving filter window size under emotions of happiness or anger.

\section{Methods}
The proposed emotion identification method based on three-axis acceleration sensor and gravity sensor embedded in mobile phone comprises the following three steps: 1) collecting and pre-processing the sensor data from subjects, 2) extracting features, and 3) training classifiers.
At the last step,
we train several classification models to evaluate and analyze their performance.

\subsection{Participants}
%【这部分关于被试招募，以及被试基本情况介绍的，可以参照相类似的文章的写法。】
To investigate identification of emotion in gaits patterns,
$59$ healthy young adult subjects(age , female = $32$) were recruited to participate in this study from University of Chinese Academy of Sciences(UCAS),
This study was approved by Institute of Psychology, Chinese Academy of Sciences and written informed consent was obtained from all subjects prior to their participation.
 Our project employed two SAMSUNG GALAXY S2 and one SAMSUNG Tab as platform(Android operation system, because the Android operating system is free, open-source,
easy for us to program to access accelerometer and gravity sensor value in cellphone and develop an APP on SAMSUNG Tab to record the start and end of activity time).
The cellphone had a $5$ Hz sample frequency, i.e., sensor recorded one piece of data per $200$ms.

Subjects were required to wear respectively one cellphone on wrist and another on ankle,
then reported his/her current emotion,
rating own anger score, and standing at specified starting line,
then performed daily walking in a fixed rectangle-shaped area for about two minutes,
then continued walking for about one minute after watching infuriating video for emotion priming.
At least three hours later(if interval is too small,
it influences participant's next emotion arousal after video),
participant was allowed to conduct the second-round experiment,
similarly, reported current emotion, rating own happiness score, then wore one cellphones on wrist and another on ankle as same as the first round,
walked for about two minutes in above area, then continued to walk for about one minute after watching prepared funny video.
The start and end of each walking time was recorded in SAMSUNG Tab APP by hand by our host.
Then we used the time series data in Tab to cut and aggregate database file data in cellphone into samples.

\subsection{Data Preprocessing}

\begin{figure}[htbp]
\centering
\centerline{\includegraphics[height=5cm,width=12cm]{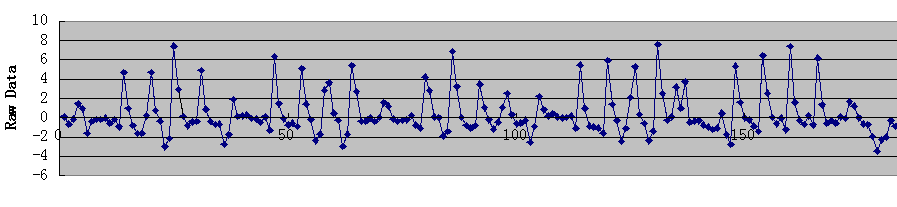}}
\caption{one raw accelerometer data of X-axis from ankle }
\label{raw-ankle}
\end{figure}

We have acquired two groups of sensor data, one is for wrist, and another is for ankle.
Each group also includes accelerometer data sets SensorLa and gravity data sets SensorGra.
According to Tab's time record, we cut every participant's walking data before and after video.
Then after SensorLa sets subtracts SensorGra sets at same time, what we get is pure accelerometer data.
Because of noise and burrs existing in data, we need to do some preprocess to pure accelerometer data,
moving average filter is suitable for time domain signal. Moving average filter expression shows as below:

\centerline{Output[i]=$\frac{1}{w}\sum_{j=0}^{w-1}Input[i+j]$}
w is adjustable for size of average process once, we set w $3$ and $5$ respectively in our data procession.
As shown in Figure.~\ref{raw-ankle}, a raw data of X-axis from ankle.

The $w$ value has significantly influence on smooth performance.
The below figures show the ankle wave signal with respect to $w$.

\begin{figure}[htbp]
\centering
\centerline{\includegraphics[height=5cm,width=12cm]{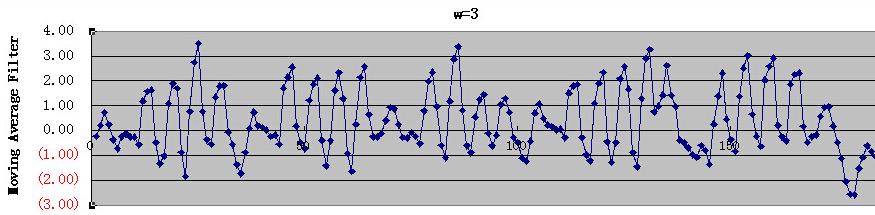}}
\caption{one raw accelerometer data of X-axis from ankle is processed by moving average filter when w is 3 }
\label{raw-ankle-3}
\end{figure}

When $w=3$,
the undulating signal has become smoother than raw data shown in Figure.~\ref{raw-ankle}.
If $w=5$, the signal become more smoother than that when w is $3$, as shown in Figure.~\ref{raw-ankle-3}.

\begin{figure}[htbp]
\centering
\centerline{\includegraphics[height=5cm,width=12cm]{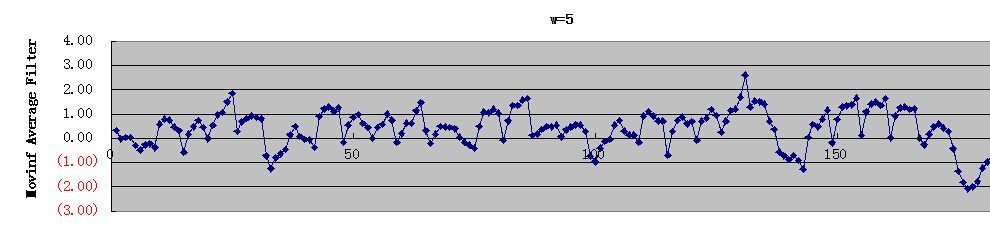}}
\caption{one raw accelerometer data of X-axis from ankle is processed by moving average filter when w is 5}
\end{figure}

But if $w$ value is too great,
it will eliminate existing minor changes in wave signal.
Though it makes wave smoother,
perhaps we lose key undulatory information of signal.
Here we respectively set w $3$ and $5$ to analyze results.

\begin{figure}[htbp]
\centering
\centerline{\includegraphics[height=5cm,width=12cm]{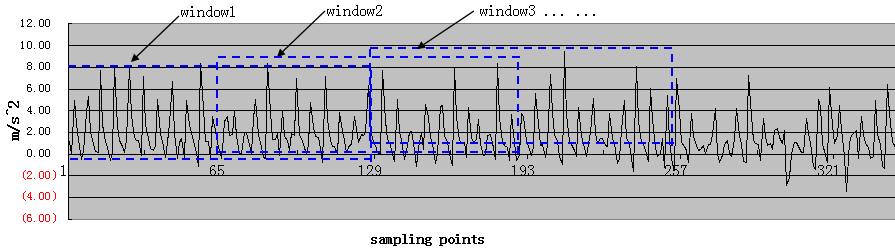}}
\caption{slice sliding window to cut accelerometer signal}
\label{signal-window}
\end{figure}

Due to the sampling frequency is $5$Hz,
i.e., the APP can access accelerometer data five times per second and write it into database one by one.
A few minutes can accumulate hundreds of pieces of records,
these records are too much to deal directly and extract features hastily.
Generally, slice sliding window is common means used to cut data into sheets or slices,
slice sliding window size we choose is $128$,
and the coverage ratio is $50\%$, like Figure.~\ref{signal-window} shown.

\subsection{Feature Extraction}
Participant has a great difference in walking with own complex behavioral traits,
different speed, and every participant's records is not same,
here we extract time domain, frequency domain, power and distribution features from each activity data sheet, including each axis's standard deviation,
means, kurtosis, skewness,
the front $32$ amplitude coefficients of FFT(the first value is current component, i.e., the axis's means),
the means and standard deviation of power spectral density(PSD), and the correlation coefficient of every two axes,
total $38$ dimensions one axis. So one activity data sheet produces a $1*(38*3)$ feature matrix.
Similarly, all participants' activity sheets produce feature matrix,
then we aggregate all these feature matrices into one matrix.

Different features show activity's properties,
means represents signal direct-current component in activity data sheet,
belonging to low frequency property. Standard deviation shows the stability of acceleration signal,
and represents degree of concentration of the motion data. because of window width is $128$,
the front $32$ amplitude coefficients of FFT denote activity property in low-frequency domain. And PSD feature explains acceleration signal from the point of energy.

All of the above features almost have big numerical difference,
even some important features with small values are ignored in classification,
and seriously affect the results.
To avoid it,
we need do Z-score normalization to features sets matrix.

In general, high dimension of feature vector increases computational complexity,
and there exists much redundant information. In order to reduce the dimension of feature vectors, and to obtain the best description of the different behavior and the best classification characteristics, dimension reduction is an essential step.

\section{Results}
For train sets we get from wrist and ankle with different w(w=$3$ and w=$5$) in the two rounds of experiment,
for the first-round,
we labeled each sample with ``natural'' or ``anger'' after PCA,
then we trained models in Weka.
Similarly, for train sets got in the second-round,
we labeled each with ``neutral'' or ``happy'', then trained models.

\subsection{Anger Emotion Identification}
In first-round experiment,
we acquired accelerometer data from wrist and ankle.
After a series of procession,
we utilized diverse kinds of classification algorithms to train models in Weka with default parameters and standard $10$-fold cross validation, including SVM, Decision Tree(J48), Random Tree, Multilayerperception and Random Forest.
Decision Tree model is explained easily and fast, but Random Forest can inspect impact between features. while SVM not just outstrips Multilayerperception in linearity and nonlinearity, but has a good performance to deal with high dimension.   \\
%【这里最好列出使用的哪些算法，加上参考文献，也可以稍微说明下选择这些算法的考虑】
The results of identification with $w=3$ from wrist and ankle are shown in Table.~\ref{classification-accuracy}.

\begin{table}[htbp]
\caption{The classification accuracy in different models when w=$3$ }
\label{classification-accuracy}
\centering
\begin{tabular}{clllll}
    \toprule
    Joint & SVM & DT & RF & MLP & RT\\
    \midrule
    wrist & $90.03\%$ & $56.25\%$ & $62.21\%$  & $63.92\%$ & $58.81\%$ \\
    ankle & $90.31\%$ & $71.31\%$ & $64.49\%$ & $59.38\%$ & $59.94\%$ \\
    \bottomrule
\end{tabular}
\begin{tablenotes}
\item [1] DT : Decision Tree.
\item [2] RF : Random Forest.
\item [3] MLP : Multilayerperception .
\item [4] RT : Random Tree.
\end{tablenotes}
\end{table}

From the above reults,
the emotion priming by watching video clips really works,
participants reported emotion arouses according to own emotion score change and had a significant influence on her or his gait.
In addition,
both wrist and ankle have a relatively higher accuracy in SVM than other models.
Meanwhile, the identification accuracy from ankle is higher than wrist.
A major reason is that the activity of hand is more complex than ankle when walking,
causing much noise which is not easily filtered out from data.

In fact,
when we set $w=5$,
the results we obtained are of greatly significant difference,
as shown in below Table~\ref{accuracy-w5}.

\begin{table}[htbp]
\caption{The classification accuracy in different models when w=$5$ }
\label{accuracy-w5}
\centering
\begin{tabular}{clllll}
    \toprule
    Joint & SVM & DT & RF & MLP & RT\\
    \midrule
    wrist & $84.61\%$ & $54.99\%$ & $59.54\%$  & $58.97\%$ & $52.99\%$ \\
    ankle & $87.46\%$ & $74.07\%$ & $65.81\%$  & $-$ & $62.68\%$  \\
    \bottomrule
\end{tabular}
\begin{tablenotes}
\item [1] - : invalid.
\end{tablenotes}
\end{table}

The results show that when w is $5$,
the evaluation results of most above models have a little higher accuracy than the results of with $ w=3$ except for SVM.
The accuracy of wrist is still lower than ankle, same as results when w is $3$.  %【这句话说的不通顺】

\subsection{Happiness Emotion Identification}
In second-round experiment,
the way we obtained accelerometer data from wrist and ankle is the same as we did in the first-round experiment.
After data preprocessing,
we run several classification algorithms to train models in Weka.
The classification results is shown in Table~\ref{tab-w3} with $w=3$.

\begin{table}[htbp]
\caption{The classification accuracy in different models when w=$3$ }
\label{tab-w3}
\centering
\begin{tabular}{clllll}
    \toprule
    Joint & SVM & DT & RF & MLP & RT\\
    \midrule
    wrist & $89.76\%$ & $61.19\%$ & $61.49\%$ & $58.51\%$ & $61.19\%$ \\
    ankle & $87.65\%$ & $74.93\%$ & $67.46\%$ & $61.19\%$ & $62.39\%$ \\
    \bottomrule
\end{tabular}
\end{table}

From above results,
we can find that the funny video arouse participants' emotion so that their gaits have a significant difference,
which makes it easy to differentiate the gaits before and after emotion priming.
Just as shown in Table~\ref{classification-accuracy},
ankle performs better to identify emotion than wrist in all above models.
The ankle accuracy reaches $87.65\%$.
Similarly, w has a great influence on classification accuracy in second-round experiment,
as shown in Table~\ref{tab4-w5}.

\begin{table}[htbp]
\caption{The classification accuracy in different models when w=$5$ }
\label{tab4-w5}
\centering
\begin{tabular}{clllll}
    \toprule
    Joint & SVM & DT & RF & MLP & RT\\
    \midrule
    wrist & $83.73\%$ & $63.88\%$ & $58.20\%$ &  $51.94\%$ & $62.69\%$ \\
    ankle & $87.65\%$ & $85.07\%$ & $70.45\%$ &  $54.32\%$ & $60.60\%$ \\
    \bottomrule
\end{tabular}
\end{table}
The table fully demonstrates that w does influence emotion identification to some extend,
and SVM has the best accuracy than other models we use, up to $87.65\%$.

\subsection{Emotions Identification}
We aggregated data sets after emotion priming in both first-round experiment and second-round experiment,
and respectively labelled them as ``anger'' and ``happy''.
The accuracy of classification is shown in Table~\ref{tab5-w3} and Table~\ref{tab6-w5}.

\begin{table}[htbp]
\caption{Anger-happy classification accuracy in different models when w=$3$ }
\label{tab5-w3}
\centering
\begin{tabular}{cllll}
    \toprule
    Joint & SVM & DT & RF  & RT\\
    \midrule
    wrist & $76.83\%$ & $63.34\%$  & $-$ &  $-$ \\
    ankle & $78.00\%$ & $74.49\%$ & $63.64\%$ & $56.60\%$ \\
    \bottomrule
\end{tabular}
\end{table}

\begin{table}[htbp]
\caption{Anger-happy classification accuracy in different models when w=$5$ }
\label{tab6-w5}
\centering
\begin{tabular}{cllll}
    \toprule
    Joint &SVM & DT & RF & RT\\
    \midrule
    wrist &$65.98\%$ & $63.05\%$ & $54.25\%$ & $54.25\%$ \\
    ankle &$87.10\%$ & $85.34\%$ & $67.16\%$ & $66.86\%$  \\
    \bottomrule
\end{tabular}
\end{table}
Form above two tables,
It is obvious that there exist significant difference among person's gaits under different emotions, e.g., anger and happy.
SVM always has better performance for ankle,
reaching $87.10\%$ to identify anger or happy emotion when w is $5$ than accuracy when w is $3$.%【这种比较可能不大合适，因为w不同，应该是比较w相同情况下不同的差别。】??

In below Table~\ref{tab7-w3hand}, Table~\ref{tab8-w3-foot} and Table~\ref{tab9-w5-hand}, anger-neutral-happy emotion confusion matrix for SVM shows that neutral emotion is easiest to be identified. When w=$5$, Table~\ref{tab10-w5-foot} shows that anger is easiest to be identified.
\begin{table}[htbp]
\caption{Anger-neutral-happy confusion matrix when w=$3$ for wrist}
\label{tab7-w3hand}
\centering
\begin{tabular}{cllll}
    \toprule
    affect &anger & neutral & happy & acc\\
    \midrule
    anger &$136$ & $7$ & $32$ & $78\%$ \\
    neutral &$18$ & $151$ & $7$ & $86\%$  \\
    happy &$43$ & $8$ & $115$ & $69\%$  \\
    \bottomrule
\end{tabular}
\end{table}

\begin{table}[htbp]
\caption{Anger-neutral-happy confusion matrix when w=$3$ for ankle}
\label{tab8-w3-foot}
\centering
\begin{tabular}{cllll}
    \toprule
    affect &anger & neutral & happy & acc\\
    \midrule
    anger &$126$ & $10$ & $39$ & $72\%$ \\
    neutral &$10$ & $152$ & $14$ & $86\%$  \\
    happy &$31$ & $10$ & $125$ & $75\%$  \\
    \bottomrule
\end{tabular}
\end{table}

\begin{table}[htbp]
\caption{Anger-neutral-happy confusion matrix when w=$5$ for wrist}
\label{tab9-w5-hand}
\centering
\begin{tabular}{cllll}
    \toprule
    affect &anger & neutral & happy & acc\\
    \midrule
    anger &$121$ & $18$ & $36$ & $69\%$ \\
    neutral &$25$ & $135$ & $16$ & $77\%$  \\
    happy &$57$ & $17$ & $92$ & $55\%$  \\
    \bottomrule
\end{tabular}
\end{table}

\begin{table}[htbp]
\caption{Anger-neutral-happy confusion matrix when w=$5$ for ankle}
\label{tab10-w5-foot}
\centering
\begin{tabular}{cllll}
    \toprule
    affect &anger & neutral & happy & acc\\
    \midrule
    anger &$148$ & $5$ & $22$ & $85\%$ \\
    neutral &$18$ & $137$ & $21$ & $78\%$  \\
    happy &$17$ & $20$ & $129$ & $78\%$  \\
    \bottomrule
\end{tabular}
\end{table}
Four confusion matrixes show that the affective state happy is easier to be misclassified as anger, similarly, anger is also easier to be misclassified as happy.
\section{Discussion}
In this paper,
we use PCA for feature selection,
and try different machine learning algorithms in Weka on our dataset.
We have tried to build models with different parameter settings,
and run $10$-fold cross validation for evaluation.

We extract $114$ features from accelerometer data,
then train models to identify person's emotion based on participant's gait.
The experimental results presented above are quite interesting and promising,
which demonstrates that there exists significant difference of walking under different emotion.
Different w values(moving average filter window size) have an evident effect on the accuracy of identification.
We find that while w becomes greater,
the sequence is smoother in time-domain.
But if w is too great,
it may ignore any tiny changes which may decrease the performance of the models.
Otherwise, small w loses an evident moving smooth performance.
When w is $3$ or $5$,
ankle has a better performance for emotion identification than wrist,
with the accuracy of $90.31\%$ in first-round experiment and $89.76\%$ in second-round experiment.
We suspect that wrist has complex additional movement when people walk. Besides, two kinds of emotion(anger-happiness) is relatively easy to be identified, whose accuracy reaches $87.10\%$.

M. Karg and R. Jenke\cite{karg2009two} presented two-fold PCA algorithm
to make a four-emotion classification: angry, happy, neutral and sad.
Their results indicate that the accuracy of angry prediction reaches $69\%$ and $76\%$ on happy by using Naive Bayes.
In our work, we find SVM works the best, reaching $90.31\%$
For our experiment, because we obtain person's pure wrist and ankle accelerometer data,
there existing no much noise information relatively,
and we analyze all features which represents one person's gait characteristics to certain extend, the results are more credible.

\section{Conclusion and Outlook}
This paper investigates identification of human emotion by natural walking.
To do so, we obtained motion data by using wrist and ankle accelerators.
After data preprocess, we extract $114$ features from each data sheet.
In four learning models we obtained in Weka with default parameters and standard $10$-fold cross validation, SVM, Decision Tree, Random Tree and Random Forest, SVM classifier has a best performance to identify personal affect.

The results of different trained models lead to the conclusion that ankle data is more capable to reveal human emotion than wrist,
best accuracy reaching $90.31\%$,
and preprocess is also key factor to improve model performance.
It is also concluded that different emotion is recognizable from human's characteristics of walking, which reaches an accuracy of $87.10\%$.

However, further consideration and improvement is required.
This includes what effect the size of slice sliding window has on identification, whether it can also improve model's performance. Besides, moving average filter(w=$3$ or $5$)used in this paper causes overlapping every two values between two adjacent average filter windows.

\section*{Acknowledgments}
The authors gratefully acknowledges the generous support from
National High-tech R\&D Program of China (2013AA01A606),
National Basic Research Program of China(2014CB744600),
Key Research Program of Chinese Academy of Sciences(CAS)(KJZD-EWL04),
and CAS Strategic Priority Research Program (XDA06030800).

\section*{References}
\bibliography{reference}
\bibliographystyle{plain}
%\bibliographystyle{elsarticle-num}
%%%%%%%%%%%%%  http://blog.sina.com.cn/s/blog_62b5588201013bah.html
\end{document}